\documentclass[letterpaper]{article} 
\usepackage{aaai25}  
\usepackage{times}  
\usepackage{helvet}  
\usepackage{courier}  
\usepackage[hyphens]{url}  
\usepackage{graphicx} 
\usepackage{natbib}  
\usepackage{caption} 
\usepackage{graphicx}
\usepackage{url} 
\usepackage{amssymb}
\usepackage{amsmath}
\usepackage{xcolor}
\usepackage{multirow}
\usepackage{algorithm}
\usepackage[algo2e,ruled,vlined]{algorithm2e}
    
    \SetCommentSty{mycommfont}

\newcommand{\sectref}[1]{Section~\ref{#1}}

\newcommand{\figref}[1]{Figure~\ref{#1}}
\newcommand{\tabref}[1]{Table~\ref{#1}}

\newcommand{\defref}[1]{Definition~\ref{#1}}
\newcommand{\agref}[1]{Algorithm~\ref{#1}}

\newtheorem{theorem}{Theorem}
\makeatletter
\@addtoreset{theorem}{section}
\makeatother

\newtheorem{definition}{Definition}

\newcounter{exampcount}
\def\land{{\wedge}}

\newcommand{\satstrong}{\models_s}
\newcommand{\notsatstrong}{\not\models_s}
\newcommand{\satweak}{\models_w}
\newcommand{\notsatweak}{\not\models_w}

\def\rmdef{\stackrel{\mbox{\rm {\small def}}}{=}}

\title{Quantitative Predictive Monitoring and Control \\ for Safe Human-Machine Interaction}

\author {
    Shuyang Dong\textsuperscript{\rm 1},
    Meiyi Ma\textsuperscript{\rm 2},
    Josephine Lamp\textsuperscript{\rm 3},
    Sebastian Elbaum\textsuperscript{\rm 1},
    Matthew B. Dwyer\textsuperscript{\rm 1},
    Lu Feng\textsuperscript{\rm 1}
}
\affiliations {
    \textsuperscript{\rm 1}University of Virginia,
    \textsuperscript{\rm 2}Vanderbilt University,
    \textsuperscript{\rm 3}DexCom \\
    \{sd3mn, se4ja, md3cn, lf9u\}@virginia.edu, meiyi.ma@vanderbilt.edu, josephine.lamp@dexcom.com
}

\begin{document}

\maketitle

\begin{abstract}
There is a growing trend toward AI systems interacting with humans to revolutionize a range of application domains such as healthcare and transportation. However, unsafe human-machine interaction can lead to catastrophic failures. We propose a novel approach that predicts future states by accounting for the uncertainty of human interaction, monitors whether predictions satisfy or violate safety requirements, and adapts control actions based on the predictive monitoring results. Specifically, we develop a new quantitative predictive monitor based on \emph{Signal Temporal Logic with Uncertainty} (STL-U) to compute a \textit{robustness degree interval}, which indicates the extent to which a sequence of uncertain predictions satisfies or violates an STL-U requirement. We also develop a new loss function to guide the uncertainty calibration of Bayesian deep learning and a new adaptive control method, both of which leverage STL-U quantitative predictive monitoring results. We apply the proposed approach to two case studies: Type 1 Diabetes management and semi-autonomous driving. Experiments show that the proposed approach improves safety and effectiveness in both case studies.
\end{abstract}

\section{Introduction} \label{sec:intro} 
There is a growing trend toward AI systems interacting with humans to revolutionize a range of application domains such as healthcare and transportation. However, unsafe human-machine interaction can lead to catastrophic failures (e.g., crashes of automated vehicles~\cite{banks2018driver} and robot-caused fatalities~\cite{yang2022robot}. Ensuring the safety of human-machine interaction poses significant challenges. First, safety is an emergent property that requires holistic reasoning about AI systems and human operators~\cite{ma2021toward}. Second, modeling human-machine interaction should account for the inherent uncertainty of human behavior. Moreover, safe and prompt decision-making under uncertainty entails predictive monitoring, i.e., making predictions about future states and monitoring if safety requirements would be violated. We concretize these challenges using a motivating example below.

Type 1 Diabetes (T1D) is a chronic disease that affects millions of patients whose pancreas produces little to no insulin to regulate blood glucose (BG). Uncontrolled diabetes may cause hypoglycemia (BG below 70 mg/dL) or hyperglycemia (BG above 180 mg/dL), which over time can lead to serious damage to organs such as the kidneys and heart. Over the past decades, advanced technologies such as continuous glucose monitoring (CGM) sensors and insulin pumps have been developed to reduce the need for patients to check BG via finger-pricking and self-injections of insulin. Recently, there have been promising breakthroughs in the development of Artificial Pancreas Systems (APS), which are automated or semi-automated closed-loop insulin delivery systems to regulate BG levels. A typical APS controller calculates insulin dosages based on CGM sensor readings and user input (e.g., meal carbohydrates). There has been increasing interest in using machine learning techniques to predict future BG levels, which are then fed into an APS controller. To guarantee a safety requirement, such as ``BG levels should be regulated within the range of 70-180 mg/dL to avoid hypoglycemia or hyperglycemia'', it is not sufficient to only check APS control actions. Instead, it is necessary to reason about the behavior of the entire closed-loop system of APS, including the insulin pump, CGM sensor, as well as patient physiology (e.g., glucose metabolism) and behavior (e.g., eating).

\begin{figure}[t]
    \centering
    \includegraphics[width=\columnwidth]{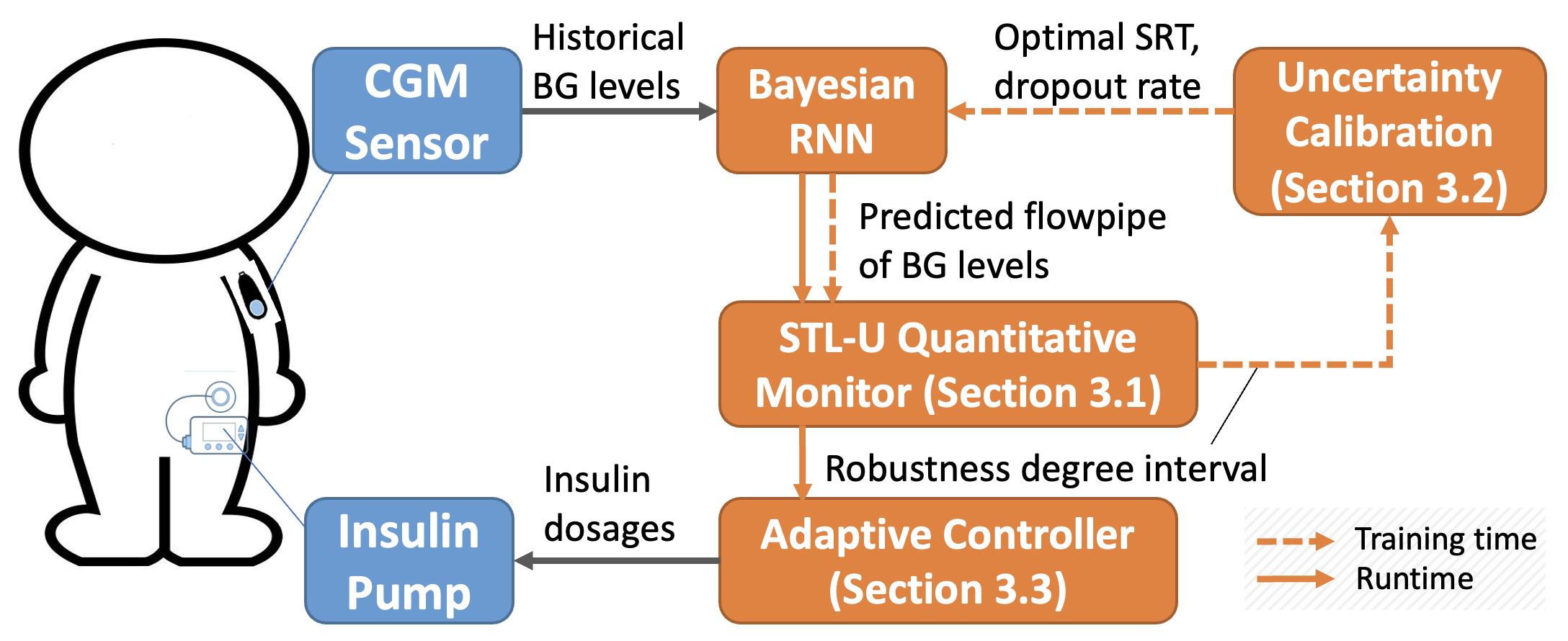}
    \caption{Proposed approach applied to T1D management.}
    \label{fig:overview}
\end{figure}

To tackle these challenges, we propose a novel logic-based quantitative predictive monitoring and control approach, as illustrated in \figref{fig:overview}. We adopt Recurrent Neural Networks (RNNs), which are well-suited to time series data~\cite{bengio2017deep}, for making sequential predictions about future BG levels. Commonly used RNNs, such as long short-term memory (LSTM) networks, are deterministic models that generate the same predictions with the same input~\cite{hochreiter1997long,ma2020stlnet}. To account for the uncertainty of human physiology and behavior, we cast deterministic RNNs into Bayesian RNNs using stochastic regularization techniques (SRTs)~\cite{gal2016uncertainty}. Bayesian RNNs yield uncertain sequential predictions with the uncertainty estimated by a sequence of posterior probability distributions.

\cite{ma2021predictive} characterize Bayesian RNN predictions as a \emph{flowpipe} signal containing an infinite set of predicted traces, and present \emph{Signal Temporal Logic with Uncertainty} (STL-U) to check if a flowpipe strongly or weakly satisfies a requirement (i.e., whether the requirement satisfaction holds for all or some traces contained in a flowpipe). Nevertheless, existing STL-U monitors do not provide quantitative information about the degree to which a flowpipe satisfies or violates a requirement, which is imperative for fine-grained decision-making in safe human-machine interactions such as diabetes management.

In this work, we develop a new STL-U quantitative monitor that computes a \emph{robustness degree interval}, indicating the degree to which a requirement is satisfied or violated. The lower and upper bounds of a robustness degree interval correspond to the worst-case and best-case estimates of the degree to which a flowpipe satisfies a requirement.
Additionally, we define a loss function that leverages STL-U quantitative monitoring results to calibrate the uncertainty estimation of Bayesian RNNs during training. We select the optimal combination of SRT and dropout rate that yields the smallest loss. Predictions made by Bayesian RNNs using the calibrated SRT and dropout rate improve the quality of uncertainty estimates with respect to requirement satisfaction.
Furthermore, we adapt control actions based on STL-U robustness degree intervals. For instance, we present a proof-of-concept adaptive APS controller that increases or decreases insulin dosages depending on the robustness degree for predicted BG levels that violate the safety requirement.
Finally, we evaluate the proposed approach through experiments using the state-of-the-art UVA/PADOVA T1D patient simulator~\cite{man2014uva}. To demonstrate the generalizability of the proposed approach, we also apply it to a second case study of semi-autonomous driving using the CARLA simulator~\cite{CARLA}, which is included in the appendix due to page limits.

\section{Background} \label{sec:background} 
\subsection{Uncertain Sequential Prediction}
Stochastic regularization techniques (SRTs) are commonly employed to transform deterministic deep learning models into Bayesian models, enabling uncertainty estimation.
In this work, we transform a deterministic RNN model into a Bayesian RNN model via SRTs. We consider four commonly used SRTs: Bernoulli dropout, Bernoulli dropConnect, Gaussian dropout, and Gaussian dropConnect~\cite{gal2016uncertainty}.
These SRTs have various ways to determine which neuron connections to drop based on sampling from a probability distribution with certain dropout rate $p$. The larger the value of $p$, the more neuron connections are retained.

A Bayesian RNN model yields a set of sequential predictions by applying Monte Carlo sampling $N$ times.
At each step $t$ along the sequence, a Gaussian distribution $\Phi_t\sim {N}(\theta_t,\sigma^2_t)$ can be estimated, whose mean $\theta_t$ and variance $\sigma_t$ are calculated based on the Monte Carlo samples $\{x^{(1)}_t, \cdots, x^{(N)}_t\}$.
The uncertainty estimates of Bayesian RNN predictions are bounded by the Gaussian distribution's confidence interval $[\Phi_t^-(\varepsilon), \Phi_t^+(\varepsilon)]$ under a confidence level $\varepsilon \in (0,1)$.
The larger the confidence interval range, the higher the estimated uncertainty.

\subsection{Signal Temporal Logic with Uncertainty}

\begin{figure}[b]
    \centering
    \includegraphics[width=\columnwidth]{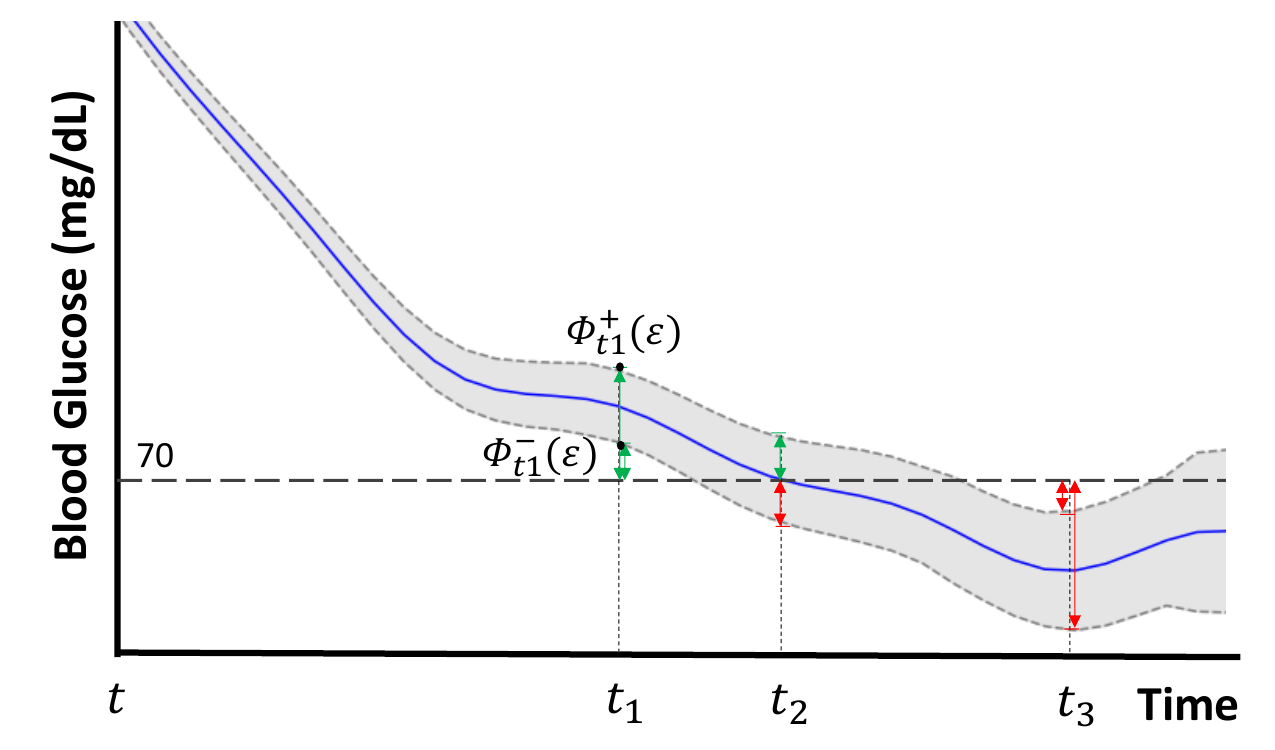}
    \caption{An example flowpipe of predicted BG levels under a confidence level $\varepsilon$.}
    \label{fig:flowpipe}
\end{figure}

\cite{ma2021predictive} characterize Bayesian RNN predictions as a \emph{flowpipe} signal $\omega$ over a discrete time domain $\mathbb{T}$. 
At each time $t$, the flowpipe contains all values bounded within a confidence interval $[\Phi_t^-(\varepsilon), \Phi_t^+(\varepsilon)]$. 
\figref{fig:flowpipe} shows an example flowpipe of predicted BG levels. 

\emph{Signal Temporal Logic with Uncertainty} (STL-U) with the following syntax is proposed in~\cite{ma2021predictive}. 
$$
\varphi := \mu(\varepsilon) \ |\ \neg \varphi \ |\ \varphi_1 \land \varphi_2 \ |\ \square_I \varphi\ |\ \lozenge_I \varphi \ |\ \varphi_1 \ {U}_I \ \varphi_2
$$
where $\square_I$, $\lozenge_I$ and ${U}_I$ are temporal operators ``always'', ``eventually'', and ``until'' with a time interval $I\subseteq \mathbb{T}$, respectively.
$\mu(\varepsilon)$ is an atomic predicate whose value is determined by a function $f(x) > 0$ for 
$x \in [\Phi_t^-(\varepsilon), \Phi_t^+(\varepsilon)]$ under a confidence level $\varepsilon$.
For example, $\Box_{[0,3]} ({BG}_{\varepsilon=90\%} > 70)$ is an STL-U formula expressing the requirement that ``the predicted BG level under a 90\% confidence level should always be above 70 mg/dL in the next three hours''.

STL-U semantics defined in~\cite{ma2021predictive} include two indices:
\emph{strong satisfaction} (i.e., all values bounded within the flowpipe's confidence interval satisfy $\varphi$), 
and \emph{weak satisfaction} (i.e., there exists some value within the flowpipe's confidence interval satisfying $\varphi$).
For example, the flowpipe shown in \figref{fig:flowpipe}
strongly satisfies $\Box_{[t,t_1]} ({BG}_{\varepsilon} > 70$), 
weakly satisfies $\Box_{[t,t_2]} ({BG}_{\varepsilon} > 70)$, 
and strongly violates $\Box_{[t,t_3]} ({BG}_{\varepsilon} > 70)$.

Additionally, a loss function, denoted by ${L}_{sat}$, is proposed in~\cite{ma2021predictive} based on STL-U strong/weak satisfaction relations to calibrate the uncertainty estimation of Bayesian RNNs by guiding the choice of SRTs and dropout rates.

\section{Approach} \label{sec:approach} 
The goal of our approach is to provide quantitative information about the degree to which an STL-U formula is satisfied or violated, which is imperative for fine-grained decision making in safe human-machine interaction. 
To achieve this goal, we develop a new STL-U quantitative monitor in \sectref{sec:quant}. The monitoring results are leveraged to improve the uncertainty calibration through a new loss function in \sectref{sec:loss} and an adaptive control method in \sectref{sec:controller}. 

\subsection{STL-U Quantitative Monitor} \label{sec:quant}

We propose a new quantitative semantics of STL-U, which computes a \emph{robustness degree} function $\rho(\varphi, \omega, t)$ indicating how much an STL-U formula $\varphi$ is satisfied or violated by a flowpipe signal $\omega$ at time $t$. 

Let $\upsilon = [\underline{\upsilon}, \overline{\upsilon}]$ denote a real-valued interval.
We define the following three interval operations: \\
{\small
${-^*}\upsilon \rmdef [- \overline{\upsilon}, - \underline{\upsilon}]$ \\
${min^*}(\upsilon_1, \dots, \upsilon_n) \rmdef [\min(\underline{\upsilon}_1, \dots, \underline{\upsilon}_n), \min(\overline{\upsilon}_1, \dots, \overline{\upsilon}_n)]$ \\
${max^*}(\upsilon_1, \dots, \upsilon_n) \rmdef [\max(\underline{\upsilon}_1, \dots, \underline{\upsilon}_n), \max(\overline{\upsilon}_1, \dots, \overline{\upsilon}_n)]$
}

\begin{definition}[STL-U quantitative semantics]\label{def:robustness}
{\small
\begin{alignat*}{2}
    &\rho(\mu(\varepsilon), \omega, t) 
        && = [\min\bigl(f(x)\bigr), \max\bigl(f(x)\bigr)], \forall x \in [\Phi_t^-(\varepsilon), \Phi_t^+(\varepsilon)] \\
    &\rho(\neg \varphi, \omega, t) 
        && = {-^*}\rho(\varphi, \omega, t)  \\
    & \rho(\varphi_1 \land \varphi_2, \omega, t) 
        && = {min^*}\bigl( \rho(\varphi_1, \omega, t), \  \rho(\varphi_2, \omega, t) \bigr)  \\
    & \rho(\square_I \varphi, \omega, t) 
        && = \underset{t' \in (t+I)}{{min^*}} \rho(\varphi, \omega, t') \\
    & \rho(\lozenge_I \varphi , \omega, t) 
        && = \underset{t' \in (t+I)}{{max^*}} \rho(\varphi, \omega, t') \\
    & \rho(\varphi_1 {U}_I \varphi_2, \omega, t)  
        && = \underset{t' \in (t+I)}{{max^*}} \Bigl( {min^*} \bigl(\rho(\varphi_2, \omega, t'), 
            \underset{t'' \in [t, t']}{{min^*}} \rho(\varphi_1, \omega, t'') \bigr) \Bigr)
\end{alignat*}}
\end{definition}

Intuitively, a robustness degree function yields an interval whose lower/upper bounds corresponding to the worst/best cases of a flowpipe satisfying or violating an STL-U formula. 
A positive (\emph{resp.} negative) robustness value indicates the degree of satisfaction (\emph{resp.} violation).

\agref{alg:monitor} illustrates STL-U quantitative monitoring algorithm based on \defref{def:robustness}.
We can apply this algorithm recursively to monitor complex STL-U formulas with multiple levels of nesting temporal operators. 
The algorithm has a linear time complexity with respect to the length of the flowpipe, $|\omega|$. 

Here is an example of checking the flowpipe in \figref{fig:flowpipe} against 
an STL-U formula $\Box_{[t,t_3]} ({BG}_{\varepsilon} > 70)$. 
First, we check the atomic predicate at each time $\tau \in [t,t_3]$, 
and compute the robustness degree interval 
$[\min\bigl(f(x)\bigr), \max\bigl(f(x)\bigr)]$ for all 
$x \in [\Phi_\tau^-(\varepsilon), \Phi_\tau^+(\varepsilon)]$, where $f(x)=x-70$.
The flowpipe at $t_2$ is bounded by $[60, 80]$ and yields a robustness degree interval $[-10, +10]$.
The flowpipe at $t_3$ is bounded by $[40, 65]$ and its robustness degree interval is $[-30, -5]$. 
Finally, we obtain a robustness degree interval for the always operator $\Box_{[t,t_3]}$
by taking the minimal of the lower/upper bounds of the atomic predicate's robustness degree intervals
over all time steps $\tau \in [t,t_3]$. 
The resulting robustness degree interval, $[-30, -5]$, indicates that the predicted flowpipe would violate the requirement by $-30$ and $-5$ in the worst and best-case scenarios, respectively.

\begin{algorithm}[H]
\caption{STL-U quantitative monitoring algorithm} \label{alg:monitor}
\begin{flushleft}
{\small
    \SetKwProg{Fn}{Function}{:}{}
    \Fn{${Monitor}({\varphi,\omega, t})$}{
        \SetKwFor{Case}{Case}{}{}
        \Switch{$\varphi$} {
            \Case{$\mu(\varepsilon)$}{
                $\rho \leftarrow [\min\bigl(f(x)\bigr), \max\bigl(f(x)\bigr)]$, for all \\
                    $x \in [\Phi_t^-(\varepsilon), \Phi_t^+(\varepsilon)]$ \\
                \Return $\rho$
            }
            \Case{$\neg \varphi$}{
                \Return ${-^*}\bigl({Monitor}({\varphi,\omega, t})\bigr)$
            }
            \Case{$\varphi_1 \land \varphi_2$}{
                \Return ${min^*}\bigl({Monitor}({\varphi_1,\omega, t}),$ \\
                    ${Monitor}({\varphi_2,\omega, t}) \bigr)$
            }
            \Case{$\square_I \varphi$}{
                $\rho \leftarrow {Monitor}(\varphi, \omega, t)$ \\
                \For {$t'\in (t + I)$}{
                    $\rho \leftarrow {min^*}\bigl(\rho, {Monitor}(\varphi, \omega, t')\bigr)$
                }
                \Return $\rho$
            }
            \Case{$\lozenge_I \varphi$}{
                $\rho \leftarrow {Monitor}(\varphi, \omega, t)$ \\
                \For {$t'\in (t + I)$}{
                    $\rho \leftarrow {max^*}\bigl(\rho, {Monitor}(\varphi, \omega, t')\bigr)$
                }
                \Return $\rho$
            }
            \Case{$\varphi_1 \ {U}_I \ \varphi_2$}{
                $\rho \leftarrow (-\infty,-\infty)$;
                $\rho_1 \leftarrow {Monitor}({\varphi_1,\omega, t})$ \\
                \For {$t'\in (t + I)$}{
                    $\rho_2 \leftarrow {Monitor}({\varphi_2,\omega, t'})$ \\
                    \For {$t'' \in [t,t']$}{
                        $\rho_1 \leftarrow {min^*}\bigl(\rho_1, {Monitor}({\varphi_1,\omega, t''})\bigr)$ 
                    }
                    $\rho \leftarrow {max^*} \bigl(\rho, {min^*}(\rho_1, \rho_2) \bigr)$            
                }
                \Return $\rho$
            }
        }
    }
}    
\end{flushleft}
\end{algorithm}

The soundness of the proposed STL-U quantitative monitor is stated below and the proof is given in the appendix. 
\begin{theorem}
Given an STL-U formula $\varphi$ and a flowpipe $\omega$, the following properties hold.
\begin{enumerate}
    \item $\underline{\rho} > 0  \Rightarrow (\omega,t) \satstrong \varphi$ 
    \item $\underline{\rho} \le 0  \Rightarrow (\omega,t) \notsatstrong \varphi$
    \item $\overline{\rho} >0  \Rightarrow (\omega,t) \satweak \varphi$
    \item $\overline{\rho} \le 0 \Rightarrow (\omega,t) \notsatweak \varphi$
\end{enumerate}
where $\underline{\rho}$ and $\overline{\rho}$ are the lower and upper bounds of the robustness degree interval
$\rho(\varphi, \omega, t)$, and $\satstrong$ (\emph{resp.} $\satweak$) denotes the strong (\emph{resp.} weak) satisfaction relations. 
\label{thm:soundness}
\end{theorem}

\subsection{Uncertainty Calibration} \label{sec:loss}

A Bayesian RNN model may produce divergent uncertainty estimates for a model trained on identical data, depending on various choices of SRTs and dropout rates. The current practice often selects an SRT and dropout rate empirically or guided by traditional deep learning metrics such as prediction accuracy, which tend to overestimate uncertainty (i.e., the wider the flowpipe, the higher the accuracy of containing the ground truth trace).

We propose a loss function based on STL-U quantitative monitoring results to guide the choice of SRTs and dropout rate. 
The proposed loss function, denoted as ${L}_{qt}(\omega, \hat{\omega}, \varphi)$, is a linear combination of two parts: 
$\eta_r(\omega, \hat{\omega}, \varphi)$ for comparing the predicted flowpipe $\omega$ and the target trace $\hat{\omega}$ in terms of the robustness degrees of satisfying an STL-U formula $\varphi$,
and $\eta_d(\omega, \hat{\omega})$ for measuring the distance between the predicted flowpipe $\omega$ and the target trace $\hat{\omega}$. 
Formally,
\begin{equation}
\eta_r(\omega, \hat{\omega}, \varphi)=\left\{
\begin{array}{lll}
\underline{\rho}(\varphi,\omega, t) &, &\hspace{0.2cm} \hat{\omega} \satstrong \varphi\\
-\overline{\rho}(\varphi,\omega, t) &, &\hspace{0.2cm} \hat{\omega} \notsatstrong \varphi\\
\end{array}\right.
\end{equation}
where $\underline{\rho}(\varphi,\omega, t)$ and $\overline{\rho}(\varphi,\omega, t)$ are the lower and upper bounds of robustness degree intervals. \\
\begin{small}
\begin{equation}
\eta_d(\omega, \hat{\omega})=\sum_{\tau =0}^{|\hat{\omega}|} \left\{
\begin{array}{lll}
0 &, & \hspace{0.2cm}  \Phi_\tau^-(\varepsilon) \leq \hat{\omega}_\tau \leq \Phi_\tau^+(\varepsilon)\\
\Phi_\tau^-(\varepsilon)  -  \hat{\omega}_{\tau} &, &\hspace{0.2cm} \hat{\omega}_{\tau} < \Phi_\tau^-(\varepsilon)\\
\hat{\omega}_{\tau}  - \Phi_\tau^+(\varepsilon) &, & \hspace{0.2cm} \Phi_\tau^+(\varepsilon) < \hat{\omega}_{\tau}
\end{array}\right.
\end{equation}
\end{small}
where $|\hat{\omega}|$ is the length of the target trace and $\hat{\omega}_{\tau}$ is the target trace's value at time $\tau$.

The loss function is then given by \\
\begin{equation}
    {L}_{qt}(\omega, \hat{\omega}, \varphi) = 
    -\beta \cdot \eta_r(\omega, \hat{\omega}, \varphi) + (1-\beta) \cdot \eta_d(\omega, \hat{\omega}) \\
\end{equation}
where $\beta \in [0,1]$ is a real-valued coefficient indicating the relative importance of the two parts.

The goal is to calibrate the uncertainty estimation of a Bayesian RNN model by choosing an optimal combination of SRT and dropout rate that yield flowpipe predictions with minimal loss ${L}_{qt}(\omega, \hat{\omega}, \varphi)$.
Intuitively, the lower the loss, the higher the quality of predicted flowpipes, 
which achieve a higher (\emph{resp.} lower) robustness degree of satisfying (\emph{resp.} violating) an STL-U formula 
and a smaller distance from the target trace.

\subsection{Adaptive Controller} \label{sec:controller}

\begin{figure}[t]
    \centering
    \includegraphics[width=0.8\columnwidth]{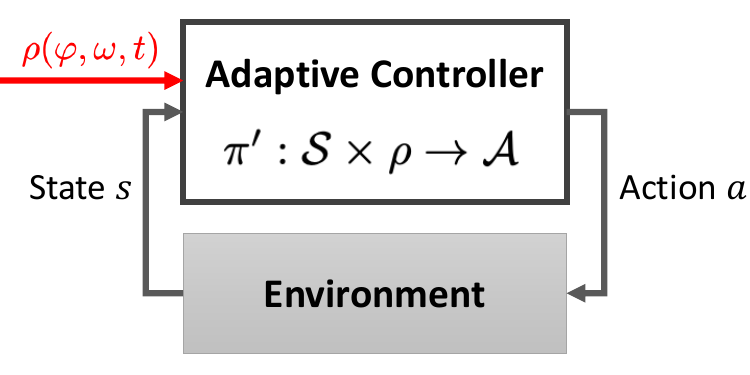}
    \caption{Adapting control actions based on STL-U quantitative monitoring results.}
    \label{fig:controller}
\end{figure}

A typical (deterministic) controller, denoted by $\pi: {S} \to {A}$, takes an input state $s \in S$ from the environment and produces an output action $a \in {A}$. 
We develop an adaptive controller $\pi': {S} \times \rho \to {A}$ that adapts control actions based on 
STL-U quantitative monitoring results $\rho(\varphi, \omega, t)$ as shown in \figref{fig:controller}.
The adaptation schema can be domain-specific. 

As a proof of concept, we present an adaptive controller based on the Basal-Bolus Controller~\cite{kovatchev2009biosimulation} included in the UVA/PADOVA T1D Simulator. 
The default Basal-Bolus Controller takes input such as the current BG level and meal carbohydrates,
and computes the basal and bolus insulin dosages as control actions to regulate BG levels.
A constant amount of basal insulin is delivered at each step, denoted by ${defaultBasal}$, 
whose value is calculated by multiplying the patient's body weight with a constant representing the steady state insulin rate per kilogram.
Additionally, the controller issues a (non-zero) bolus insulin at a step when the patient takes a meal with carbohydrates. 
The bolus insulin dosage, denoted by ${mealBolus}$, is calculated based on the amount of carbohydrates, the current and target BG levels, and the patient's carbohydrate ratio and correction factor. 

\begin{algorithm}[H]
\caption{Adapting a Basal-Bolus Controller} \label{alg:controller}
{\small
\SetKwInOut{Input}{Input}
\SetKwInOut{Output}{Output}
\SetKwFor{Case}{Case}{}{}
\Input{STL-U quantitative monitoring results $\rho(\varphi_l,\omega,t)$ and $\rho(\varphi_h,\omega,t)$, current BG level $g_t$, next planned meal time $t_m$,time-window length $K$, and ${bolusFlag}$ for a meal bolus}
\Output{Control action $a_t$ at time $t$}
\tcp{Adapting basal insulin dosages}
${basal} \leftarrow {defaultBasal}$ \\
\If{$\underline{\rho}(\varphi_l,\omega,t) < -20$}{
    ${basal} \leftarrow 0$
}
\ElseIf{$-20 \le \underline{\rho}(\varphi_l,\omega,t) \le 0$}{
    ${basal} \leftarrow {defaultBasal} \times 0.8$
}
\ElseIf{$-70 \le \underline{\rho}(\varphi_h,\omega,t) \le 0$}{
    ${basal} \leftarrow {defaultBasal} \times 1.2$
}
\ElseIf{$\underline{\rho}(\varphi_h,\omega,t) < -70$}{
    ${basal} \leftarrow {defaultBasal} \times 1.5$
}
\tcp{Adapting bolus insulin timing}
${bolus} \leftarrow 0$ \\
\If{${bolusFlag}== {false}$}{
    \If{$t_m - K \le t < t_m$}{
        \If{$\underline{\rho}(\varphi_l,\omega,t) \le 0$ or $g_t \le 70$}{
            ${bolus} \leftarrow 0$
        }\Else{
            ${bolus} \leftarrow {mealBolus}$;
            ${bolusFlag} \leftarrow {true}$
        }
    }\ElseIf{$t = t_m$}{
        ${bolus} \leftarrow {mealBolus}$;
        ${bolusFlag} \leftarrow {true}$
    }
}
\Return $a_t = \langle {basal}, {bolus}\rangle$
}
\end{algorithm}

\agref{alg:controller} adapts the Basal-Bolus Controller based on quantitative results of monitoring two STL-U formulas:
$\varphi_l = \Box_{[0,\infty)} ({BG}_{\varepsilon} > 70)$
and $\varphi_h = \Box_{[0,\infty)} ({BG}_{\varepsilon} < 180)$.
It decreases or increases basal insulin dosages based on the worst-case robustness degrees of 
predicted hypoglycemias and hyperglycemias as follows. 
It reduces the ${basal}$ dose to zero when 
$\underline{\rho}(\varphi_l,\omega,t) < -20$, 
indicating severe hypoglycemia with BG below 50 mg/dL.
It decreases the ${basal}$ dose to 80\% of the default when $-20 \le \underline{\rho}(\varphi_l,\omega,t) \le 0$, 
indicating mild hypoglycemia. 
It increases the ${basal}$ dose to 120\% when $-70 \le \underline{\rho}(\varphi_h,\omega,t) \le 0$,
indicating mild hyperglycemia.
Finally, it  increases the ${basal}$ dose to $150\%$ of the default 
when $\underline{\rho}(\varphi_h,\omega,t) < -70$,
indicating severe hyperglycemia with BG above 250 mg/dL.
We set these BG thresholds and basal change percentages following medical guidelines~\cite{care2022care}.

Furthermore, the adaptive controller adapts the time to administer a bolus insulin 
based on the current and predicted BG levels. 
Studies~\cite{slattery2018optimal} have shown that taking a bolus 15-30 minutes before a meal
helps to improve the glucose control.
The optimal bolus timing varies with patient circumstances and insulin effects.
Drawing on this insight, we design the adaptive controller to encourage pre-meal boluses. 
As shown in \agref{alg:controller}, 
the adaptive controller decides the bolus timing by 
checking the current and predicted BG levels from $K$ steps before a meal,
where $K$ is a constant recommended by medical guidelines (e.g., 45 minutes).
The adaptive controller would not issue a pre-meal bolus when the current or predicted BG levels are below 70 mg/dL
because of the risk of hypoglycemia.

\section{Experiments} \label{sec:exp} 
We evaluate the proposed approach via experiments using the UVA/PADOVA T1D Patient Simulator~\cite{man2014uva}, which has been approved by the U.S. Food and Drug Administration (FDA) for pre-clinical experiments with \emph{in silico} populations. 
We investigate three research questions:
\begin{itemize}
    \item \textbf{RQ1}: How useful is the proposed loss function for the uncertainty calibration of Bayesian RNN predictions?
    \item \textbf{RQ2}: How good is the proposed predictive monitor for the early and accurate detection of safety hazards?
    \item \textbf{RQ3}: How safe and effective is the proposed predictive monitoring and control approach in the closed-loop simulation of T1D management?
\end{itemize}

\subsubsection{Datasets and experimental setup.}
We use the simulator to generate data based on 30 virtual patient profiles including: 10 adults, 10 adolescents and 10 children. Each patient is simulated for an 85-day period, with each time step in the simulation representing 3 minutes.
We use 70-day, 5-day, and 10-day data for training, validation, and testing, respectively. 
We segment the data into samples of 20 steps (1 hour) by sliding windows, and obtain about 413,326 samples in total for each patient population. Violations are observed in 23\% of adolescent data, 48\% of adult data, and 52\% of child data.

We train three different LSTM models for adults, adolescents, and children, respectively. They seek to predict BG levels in the future 30 minutes using the historical data of the past 30 minutes. 
The data inputs include CGM sensor readings, meal carbohydrates, insulin dosages, low/high blood glucose indexes, hypo/hyper risk, and time of a day. 
Each model is trained for 50 epochs. 
Given a chosen SRT and dropout rate, we repeat Bayesian LSTM predictions for 30 times using the Monte Carlo method. 
We estimate Gaussian distributions using these predictions and obtain flowpipes under a 95\% confidence level. 
Using a higher confidence level broadens the predicted flowpipe, which can increase requirement violations. While this may appear conservative, it aligns with our focus on safety in STL-U requirements, where conservatism helps mitigate risks proactively.
We use the validation datasets to select the optimal SRT and dropout rate under the guidance of the proposed loss function, and use the selected optimal SRT and dropout rate to generate Bayesian LSTM predictions for the testing datasets. 

Lastly, we apply the proposed predictive monitoring and control approach in the closed-loop simulation of T1D management over a 7-day period (different from the 85-day data mentioned before) for 30 virtual patients.

The experiments were run on a machine with 2.1GHz CPU, Nvidia Quadro RTX5000 GPU, 128GB memory, and CentOS 7 operating system.

\subsection{RQ1: Evaluation of Uncertainty Calibration} \label{sec:rq1}

\begin{figure}[t]
    \centering
    \includegraphics[width=.9\columnwidth]{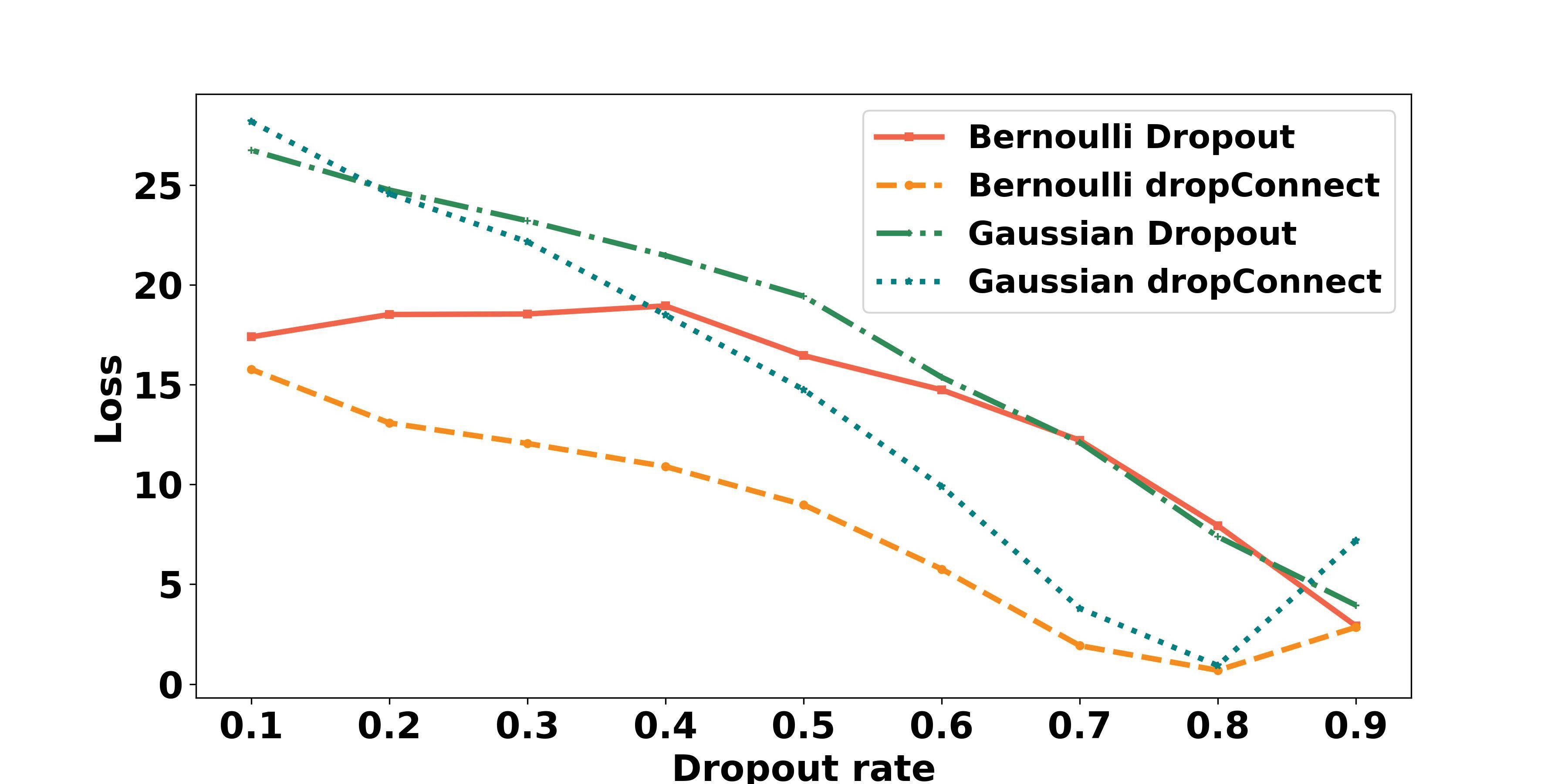}
    \caption{Loss values of using different SRTs and dropout rates in the Bayesian LSTM model for adults.}
    \label{fig:adult_loss}
\end{figure}

\figref{fig:adult_loss} plots values of the proposed loss function ${L}_{qt}$
with respect to an STL-U formula $\varphi = \Box_{[0,\infty)} (70 < {BG}_{\varepsilon} < 180)$,
when varying SRTs and dropout rates of the Bayesian LSTM model for adults with the validation dataset. 
The results show that Bernoulli dropConnect with a dropout rate of 0.8 is the best choice with the smallest loss.
For the LSTM models of adolescents and children (plots omitted due to page limits), we found Bernoulli dropConnect with a rate of 0.9 and Gaussian Dropout with a rate of 0.9 to be the best, respectively. 

We compare the performance of the proposed loss function ${L}_{qt}$ 
with two baselines taken from~\cite{ma2021predictive}:
${L}_{acc}$ which captures prediction accuracy 
(i.e., whether the target trace is entirely contained in a predicted flowpipe),
and ${L}_{sat}$ which captures STL-U strong/weak satisfaction.
We evaluate the \emph{F1 scores of requirement satisfaction} metric defined as
$\frac{{TP}}{{TP} + \frac{1}{2} ({FP} + {FN})}$, 
where ${TP}$ denotes the number of true positives 
(i.e., when the target trace satisfies $\varphi$ and the predicted flowpipe $\omega$ yields $\underline{\rho}(\varphi,\omega,t) > 0$), 
${FP}$ denotes the number of false positives 
(i.e., when the target trace violates $\varphi$ and the predicted flowpipe $\omega$ yields $\underline{\rho}(\varphi,\omega,t) > 0$),  
and ${FN}$ denotes the number of false negatives
(i.e., when the target trace satisfies $\varphi$ and the predicted flowpipe $\omega$ yields $\underline{\rho}(\varphi,\omega,t) < 0$).

\begin{table}[t]
{
\centering
\begin{tabular}{|l|c|c|c|}
\hline
\textbf{Model} &  ${L}_{acc}$ & ${L}_{sat}$ & ${L}_{qt}$ (proposed)\\
\hline
Adults($\beta=0.5$) & 0.66 & 0.88 & \textbf{0.93} \\
Adolescents($\beta=0.5$) & 0.38 & 0.60 & \textbf{0.71} \\
Children($\beta=0.5$) & 0.68 & 0.81 & \textbf{0.90} \\
\hline
\end{tabular}}
\caption{F1 scores of requirement satisfaction for comparing the proposed loss function with two baselines.}\label{tab:rq1}
\end{table}

\tabref{tab:rq1} shows F1 scores achieved by Bayesian LSTM predictions generated for the testing datasets using the optimal SRTs and dropout rates selected via different loss functions.
${L}_{acc}$ has the worst results because it tends to over-estimate the uncertainty (i.e., making the flowpipe wider) in order to improve the prediction accuracy.  
Both ${L}_{sat}$ and ${L}_{qt}$ check if the target trace and the predicated flowpipe yield consistent STL-U monitoring results. 
${L}_{qt}$ achieves higher F1 scores than ${L}_{sat}$ in all three models,
because ${L}_{qt}$ accounts for quantitative information in the form of robustness degree of requirement satisfaction.

\emph{In summary, experiments demonstrate that the proposed loss function can be used to calibrate the uncertainty estimation of Bayesian RNNs by guiding the selection of optimal SRTs and dropout rates, which improve the quality of uncertainty estimates and predictions. }

\subsection{RQ2: Evaluation of Predictive Monitors} \label{sec:rq2}

We compare the performance of the proposed STL-U quantitative predictive monitor with a baseline predictive monitor that does not account for the uncertainty. 
Specifically, we use the STL monitor~\cite{maler2004monitoring} to check the predicted flowpipes' mean traces as the baseline. 
For a fair comparison, both predictive monitors use the same Bayesian LSTM models equipped with the optimal SRTs and dropout rates.
We consider two performance metrics:
the average pre-alert time (i.e., the time interval between the earliest point when a predictive monitor detects an impending hazard and its actual occurrence time) over all possible safety hazards,
and F1 score of requirement satisfaction. Following the convention in hazard calculations for artificial pancreas systems~\cite{lum2021real}, we make the assumption that if two hazards of the same type happen closely within 30 minutes, then they are counted as one hazard.

\begin{table}[h]
{%
\centering
\begin{tabular}{|l|l|cc|cc|}
\hline
& & \multicolumn{2}{c}{\textbf{Baseline}} & \multicolumn{2}{|c|}{\textbf{Proposed}} \\
\textbf{Model} & \textbf{Hazard} & \textbf{Time} & \textbf{F1} & \textbf{Time} & \textbf{F1} \\
\hline
\multirow{3}{*}{Adults} & Hypo & 0.6  & 0.54 & \textbf{23.9} & \textbf{0.96} \\
                        & Hyper & 1.9 & 0.56 & \textbf{22.2} & \textbf{0.63} \\
                        & Overall & 1.2 & 0.54 & \textbf{23.0} & \textbf{0.93} \\
\hline
\multirow{3}{*}{Adolescents} & Hypo & 4.2 & \textbf{0.57} & \textbf{24.9} & 0.48 \\ 
                        & Hyper & 10.6 & \textbf{0.82} & \textbf{22.6} & 0.78 \\
                        & Overall & 9.2 & \textbf{0.78} & \textbf{23.1} & 0.71 \\ 
\hline
\multirow{3}{*}{Children} & Hypo & 3.9 & 0.89 & \textbf{13.1} & \textbf{0.91} \\
                        & Hyper & 21.2 & 0.71 & \textbf{27.7} & \textbf{0.75} \\
                        & Overall & 10.6 & 0.88 & \textbf{18.7} & \textbf{0.90} \\ 
\hline
\end{tabular}}
\caption{Comparing the baseline and the proposed predictive monitors in terms of average pre-alert time (minutes) and F1 scores of requirement satisfaction.}\label{tab:rq2}
\end{table}

\tabref{tab:rq2} shows that, compared to the baseline, the proposed STL-U quantitative predictive monitor leads to earlier detection (i.e., larger pre-alert time) of impending hazards across all three patient populations with statistical significance. 
The results of paired t-test are 
(t(127)=28.2, p$<$0.01, d=3.2, significant) for adults,
(t(316)=22.6, p$<$0.01, d=1.4, significant) for adolescents,
and (t(354)=17.0, p$<$0.01, d=0.7, significant) for children.  
\tabref{tab:rq2} also shows that the proposed predictive monitor achieves higher F1 scores than the baseline for 
adults and children models, indicating more accurate detection of impending hazards. 
But the proposed predictive monitor has slightly lower F1 scores than the baseline for the adolescents model. 
One possible explanation is that the simulated adolescent patients have 
higher glycemic variability index (1.96) than adults (1.69) and children (1.80),
making the prediction more challenging.  

\emph{In summary, experiments demonstrate that the proposed STL-U quantitative monitor can provide early and accurate detection of impending safety hazards.}

\subsection{RQ3: Closed-Loop Simulation} \label{sec:rq3}

We apply the proposed STL-U quantitative monitor and adaptive controller in the closed-loop simulation of T1D management. 
We use the simulator's Basal-Bolus Controller~\cite{kovatchev2009biosimulation} as a baseline for comparison. 

We measure the following commonly used metrics for the safety and effectiveness of T1D management:
\emph{number of hazards} (i.e., number of hypoglycemias and hyperglycemias that occur to a patient during the simulated period) and 
\emph{time in range} (i.e., the percentage of time that a simulated patient's BG levels stay within the range of 70-180 mg/dL).

\begin{figure}[t]
    \centering
    \includegraphics[width=\columnwidth]{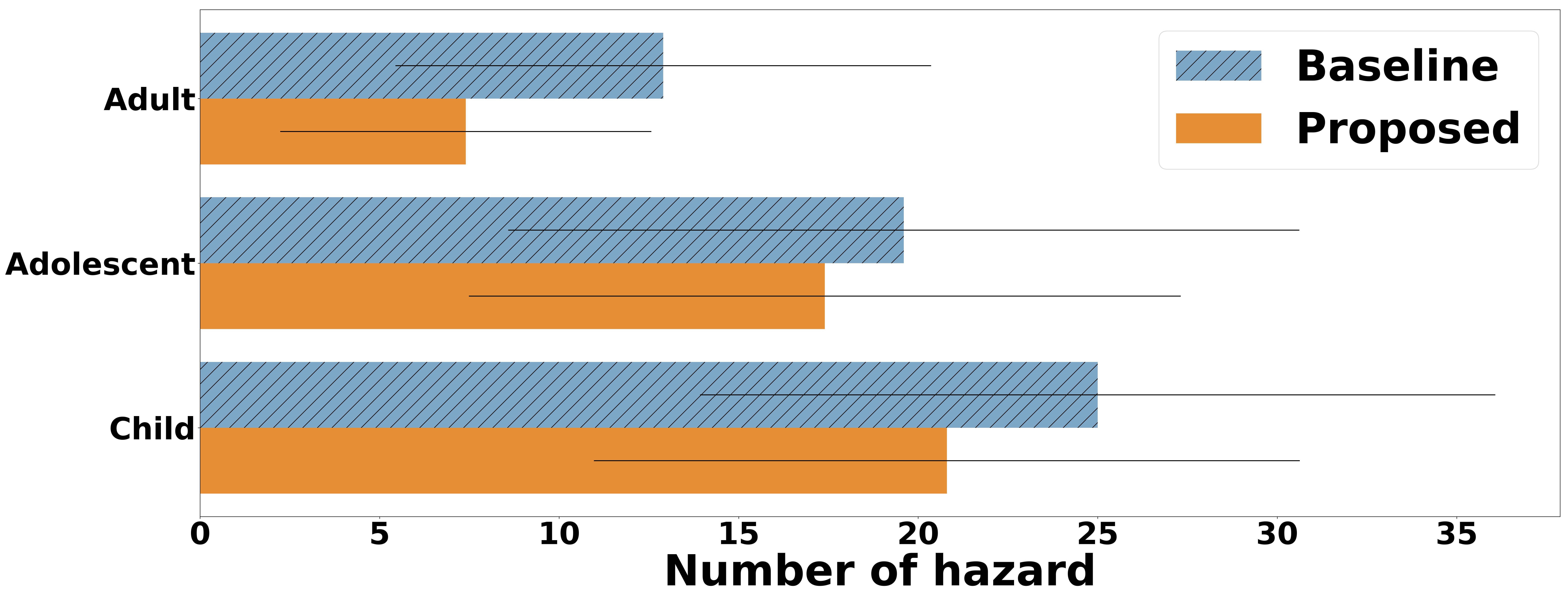}
    \caption{Comparing the number of hazards that occurred during the simulated 7-day period 
    (mean and standard deviation over each patient population).}
    \label{fig:hazards}
\end{figure}

\figref{fig:hazards} shows that the proposed approach reduces the average number of hazards for all three types of patients. 
The results of paired t-test are 
(t(9)=-3.5, p=0.01, d=-0.9, significant) for adults,
(t(9)=-1.9, p=0.09, d=-0.2, insignificant) for adolescents,
and (t(9)=-1.0, p=0.33, d=-0.4, insignificant) for children.

\begin{table}[h]
{%
\centering
\begin{tabular}{|l|l|c|c|}
\hline
\textbf{Model} & \textbf{Metric} & \textbf{Baseline} & \textbf{Proposed}  \\
\hline
\multirow{3}{*}{Adults} & Time in range & 90.9\% & \textbf{96.3\%} \\
                        & Hypo time & 4.8\% & \textbf{2.4\%} \\
                        & Hyper time & 4.3\% & \textbf{1.3\%} \\
\hline
\multirow{3}{*}{Adolescents} & Time in range & 80.9\% & \textbf{85.2\%} \\
                        & Hypo time & 4.2\% & \textbf{3.3\%} \\
                        & Hyper time & 14.9\% & \textbf{11.5\%} \\   
\hline
\multirow{3}{*}{Children} & Time in range & 63.6\% & \textbf{74.8\%} \\
                        & Hypo time & 29.4\% & \textbf{19.3\%} \\
                        & Hyper time & 7.0\% & \textbf{5.9\%} \\  
\hline
\end{tabular}}
\caption{Comparing the baseline controller and the proposed approach's performance in closed-loop simulation.}\label{tab:rq3}
\end{table}

\tabref{tab:rq3} shows that the proposed approach improves the time in range across all three patient populations with statistical significance. 
The results of paired t-test are 
(t(9)=3.4, p=0.01, d=0.6, significant) for adults,
(t(9)=2.7, p=0.02, d=0.3, significant) for adolescents,
and (t(9)=2.6, p=0.03, d=0.4, significant) for children.
We observe that adults have the most effective glucose control, 
while adolescents (\emph{resp.} children) tend to have more hyperglycemias (\emph{resp.} hypoglycemias) in the simulation.

\emph{In summary, results of the closed-loop simulation demonstrate that the proposed predictive monitoring and control approach improves the safety and effectiveness of T1D management by increasing the time in range and decreasing the number of safety hazards. }

\section{Related Work} \label{sec:related} 
\subsubsection{Safe human-machine interaction via formal methods.}
Traditional methods of human-machine interaction design primarily rely on user studies for safety assurance~\cite{sharp2019interaction}. To reduce the burden of human testing, model-based design using formal methods has been explored to provide mathematically rigorous safety guarantees for human-machine interaction~\cite{bolton2013using}. For example, a control protocol for an unmanned aerial vehicle (UAV) is synthesized by modeling the human-UAV interaction as a two-player stochastic game~\cite{feng2016synthesis}. There are also several works on the formal specification and verification of human-robot interaction~\cite{luckcuck2019formal}.

However, these existing works mostly focus on modeling and analyzing human-machine interaction at design time. By contrast, in this work, we consider the predictive monitoring and control of safe human-machine interaction at runtime. We use data-driven RNN models as abstract representations of complex human behaviors, capturing their uncertainty through Bayesian deep learning. We then develop logic-based approaches to monitor predictions made by Bayesian RNNs and adapt control actions based on the monitoring results.

\subsubsection{Logic-based predictive monitoring and control.}
Temporal logic specifications such as Signal Temporal Logic (STL) have been used for runtime monitoring (also called runtime verification) and control of cyber-physical systems, such as automobiles and medical devices~\cite{bartocci2018specification}.
Recently, there have been increasing efforts on predictive monitoring, which checks predictions about future states to support prompt decision making.
For instance, \cite{qin2020clairvoyant} apply an STL monitor to check glucose levels predicted by an ARIMA statistical model.
\cite{yoon2021predictive} presents a logic-based Bayesian intent inference to forecast a robot's future positions and avoid impending collisions.
\cite{ma2021predictive} presents an STL-U predictive monitor to check predictions of air pollution and traffic in smart cities. 

Following this line of work, we contribute to the state-of-the-art by developing a novel STL-U \emph{quantitative} predictive monitor that computes \emph{robustness degrees} indicating how far a signal is from satisfying or violating a logic specification. 
Our work is inspired by~\cite{fainekos2009robustness}, which introduces the concept of robustness degree for checking continuous-time (single-sequence) signals against temporal logic specifications. 
We adopt this concept and extend it for checking flowpipe signals that represent uncertain sequential predictions made by Bayesian RNNs. 

There is some related work on uncertainty-aware STL monitoring. For example, \cite{baharisangari2021uncertainty} define the lower and upper bounds of robustness by selecting a single trace out of an interval trajectory. \cite{visconti2021online} define lower and upper bounds for specific time points with missing values for a single trace.
\cite{lindemann2023conformal} construct prediction regions that quantify prediction uncertainty using conformal prediction, a statistical tool for uncertainty quantification. By contrast, we define STL-U quantitative semantics over all possible traces included in a flowpipe, capturing the uncertainty of Bayesian RNNs via confidence intervals at every time point.

Lastly, existing work on utilizing predictive monitoring results for downstream tasks such as control and design modification is limited. This work addresses this gap by developing a new loss function to guide the uncertainty calibration of Bayesian deep learning and a new adaptive control method, both of which leverage STL-U quantitative predictive monitoring results.

\section{Conclusion} \label{sec:conclu} 

In this work, we present a logic-based quantitative predictive monitoring and control approach to enhance the safety of human-machine interaction under uncertainty. Using Bayesian RNN models, we represent the uncertainty of human behavior and employ a novel STL-U quantitative predictive monitor to compute robustness degree intervals, which indicate the satisfaction or violation of STL-U requirements. We design a new loss function leveraging STL-U results to optimize uncertainty estimation in Bayesian RNNs by selecting the best combination of SRT and dropout rate. Additionally, adaptive controllers adjust control actions based on robustness intervals.

Experiments with a T1D patient simulator demonstrate that the proposed approach enables early and accurate detection of safety hazards and improves the safety and effectiveness of T1D management. Furthermore, the loss function outperforms state-of-the-art baselines in uncertainty estimation. Results from a semi-autonomous driving case study also show enhanced safety, confirming the approach’s generalizability.

Future work includes testing alternative RNN models, developing principled adaptive control for broader domains, incorporating priority-based dynamic enforcement of requirements, validating in real-world settings, and extending to diverse case studies like human-robot interaction.

\section*{Acknowledgments}
This work was supported in part by the U.S. Air Force Office of Scientific Research under Grant FA9550-21-1-0164 and the U.S. National Science Foundation under Grants CCF-1942836, CCF-2131511, 2220401, and 2427711. 
The opinions, findings, conclusions, or recommendations expressed in this material are those of the author(s) and do not necessarily reflect the views of the sponsoring agencies.

\appendix
\section{Correctness}

\setcounter{theorem}{0}

\begin{theorem}\label{thm:soundness}
Given an STL-U formula $\varphi$ and a flowpipe $\omega$, the following properties hold.
\begin{enumerate}
    \item $\underline{\rho} > 0  \Rightarrow (\omega,t) \satstrong \varphi$ 
    \item $\underline{\rho} \le 0  \Rightarrow (\omega,t) \notsatstrong \varphi$
    \item $\overline{\rho} >0  \Rightarrow (\omega,t) \satweak \varphi$
    \item $\overline{\rho} \le 0 \Rightarrow (\omega,t) \notsatweak \varphi$
\end{enumerate}
where $\underline{\rho}$ and $\overline{\rho}$ are the lower and upper bounds of the robustness degree interval
$\rho(\varphi, \omega, t)$, and $\satstrong$ (\emph{resp.} $\satweak$) denotes the strong (\emph{resp.} weak) satisfaction relations. 
\label{thm:soundness}
\end{theorem}
We prove the first property $\underline{\rho} > 0  \Rightarrow (\omega,t) \satstrong \varphi$ by structural induction below. The rest of the three properties can be proved similarly.
\begin{itemize}
    \item Base case $\mu(\varepsilon)$: \\
        Based on \defref{def:robustness}, $\underline{\rho}$ is the minimal value of $f(x)$ for all $x \in [\Phi_t^-(\varepsilon), \Phi_t^+(\varepsilon)]$. If $\underline{\rho} > 0$, then $f(x) > 0$ for all $x$ bounded within the flowpipe's confidence interval. Thus, the flowpipe $\omega$ strongly satisfies $\mu(\varepsilon)$ at time $t$. 
    \item Inductive case $\neg \varphi$:\\
        Since $\underline{\rho}= - \max\bigl(f(x)\bigr) > 0$ for all $x \in [\Phi_t^-(\varepsilon), \Phi_t^+(\varepsilon)]$, we have $f(x) < 0$ for all the flowpipe's values bounded within its confidence interval. Thus, the flowpipe $\omega$ strongly satisfies $\neg \varphi$ at time $t$.
    \item Inductive case $\varphi_1 \land \varphi_2$:\\
        Let $\underline{\rho}_1$ and $\underline{\rho}_2$ denote the lower bounds of robustness degree intervals
        $\rho(\varphi_1, \omega, t)$ and $\rho(\varphi_2, \omega, t)$, respectively.\\ 
        Since $\underline{\rho} = {min^*}\bigl( \underline{\rho}_1, \underline{\rho}_2 \bigr) > 0$,
        we have both $\underline{\rho}_1 >0$ and $\underline{\rho}_2 >0$. 
        Thus, the flowpipe $\omega$ strongly satisfies $\varphi_1 \land \varphi_2$ at time $t$.
    \item Inductive case $\square_I \varphi$:\\
        Since $\underline{\rho} = {{min^*}_{t' \in (t+I)}} \underline{\rho}(\varphi, \omega, t') > 0$
        , we have \\
        $\underline{\rho}(\varphi, \omega, t') > 0$ for every time step $t' \in (t+I)$. 
        Thus, the flowpipe $\omega$ strongly satisfies $\square_I \varphi$ at time $t$. 
    \item Inductive case $\lozenge_I \varphi$:\\
        Since $\underline{\rho} = {{max^*}_{t' \in (t+I)}} \underline{\rho}(\varphi, \omega, t') > 0$, we have $\underline{\rho}(\varphi, \omega, t') > 0$ for at least one step $t' \in (t+I)$.
        Thus, the flowpipe $\omega$ strongly satisfies $\lozenge_I \varphi$ at time $t$. 
    \item Inductive case $\varphi_1 \ {U}_I \ \varphi_2$:\\
        Based on \defref{def:robustness}, \\
        $\underline{\rho} = {{max^*}_{t' \in (t+I)}} 
        \Bigl( {min^*} \bigl(\underline{\rho}(\varphi_2, \omega, t'), \\ 
            {{min^*}_{t'' \in [t, t']}} 
            \underline{\rho}(\varphi_1, \omega, t'') \bigr) \Bigr)$.
          \\
        Since $\underline{\rho} > 0$, there must exist a step $t' \in (t+I)$ such that \\
        ${min^*} \bigl(\underline{\rho}(\varphi_2, \omega, t'), 
            {{min^*}_{t'' \in [t, t']}} 
            \underline{\rho}(\varphi_1, \omega, t'') \bigr) > 0$, 
        which is equivalent to \\
        $\underline{\rho}(\varphi_2, \omega, t') > 0$ and 
        $\underline{\rho}(\varphi_1, \omega, t'') > 0$ for all $t'' \in [t, t']$.
        Thus, there exists a time $t' \in (t+I)$ with $(\omega,t') \satstrong \varphi_2$
        and $(\omega,t'') \satstrong \varphi_1$ for all $t'' \in [t, t']$.
        By definition, the flowpipe $\omega$ strongly satisfies $\varphi_1 \ {U}_I \ \varphi_2$ at time $t$.
\end{itemize}


\section{Case Study: Semi-Autonomous Driving} 

To demonstrate the generalizability of the proposed approach, we apply it to a second case study of semi-autonomous driving using the CARLA simulator. We consider a benchmark scenario of car following on a straight road provided in~\cite{SafeBench}. The goal of vehicle control is to smooth the acceleration, thereby avoiding hard brakes and sharp accelerations during car following to ensure safety and comfort.
We specify the requirement using an STL-U formula:
$\varphi = \Box_{[0,\infty)} [({acceleration}_{\varepsilon} > -6.0) \land ({acceleration}_{\varepsilon} < 6.0)]$.  

\begin{algorithm}[H]
\caption{Adapting a Vehicle Controller} \label{alg:controller_driving_case}
{\small
\SetKwInOut{Input}{Input}
\SetKwInOut{Output}{Output}
\SetKwFor{Case}{Case}{}{}
\Input{STL-U quantitative monitoring results $\rho(\varphi,\omega,t)$, $\rho_{thre}$, 
    $\rho_{corr}$, ${currentSpeed}$, ${minSpeed}$, ${maxThrottle}$, ${maxBrake}$, ${currentThrottle}$, ${currentBrake}$, ${meanBrake}$ in past 5 steps, and ${meanThrottle}$ in past 5 steps}
\Output{Control action $a_t$ at time $t$}
${brake} \leftarrow {currentBrake}$,
${throttle} \leftarrow {currentThrottle}$,
${\delta} \leftarrow {1/abs(\underline{\rho}(\varphi,\omega,t)+\rho_{corr})} $\\
\If{${currentSpeed} \geq {minSpeed}$}{
    \If{$\underline{\rho}(\varphi,\omega,t) > \rho_{thre}$}
    {${brake} \leftarrow \min({currentBrake}, {maxBrake})$\\
    ${throttle} \leftarrow \min({currentThrottle}, {maxThrottle})$}
    
    \Else{
    \If{${ViolationOnDeceleration}$}
    {${brakeAdjustParam}=1+\delta$\\
    ${newBrake}= \\ {brakeAdjustParam}\times{meanBrake}$\\
    ${brake} \leftarrow \min({newBrake}, {maxBrake})$\\
    ${throttle} \leftarrow 0$}
    \ElseIf{${ViolationOnAcceleration}$}
    {${throttleAdjustParam}=1+\delta$ \\
    ${newThrottle}= \\{throttleAdjustParam}\times{meanThrottle}$ \\
    ${throttle} \leftarrow \min({newThrottle}, {maxThrottle})$\\
    ${brake} \leftarrow 0$}
    
    \ElseIf{${ViolationOnBoth}$}
    {${brakeAdjustParam}=
    \max(1-\delta, 0)$\\
    ${newBrake}= \\{brakeAdjustParam}\times{meanBrake}$\\
    ${brake} \leftarrow \min({newBrake}, {maxBrake})$\\
    ${throttle} \leftarrow 0$}
    }
}
\Else{
${throttleAdjustParam}=\max(1-\delta, 0)$\\
${newThrottle}={throttleAdjustParam}\times{meanThrottle}$ \\
    ${throttle} \leftarrow  \min({newThrottle}, {maxThrottle})$\\
     ${brake} \leftarrow 0$
}
${brake} \leftarrow \max({brake}, 0)$\\
${throttle} \leftarrow \max({throttle}, 0)$\\
\Return $a_t = \langle {brake}, {throttle}\rangle$
}
\end{algorithm}

\subsubsection{Adaptive controller.}
\agref{alg:controller_driving_case} adapts the vehicle controller based on the STL-U quantitative monitoring results. 
Specifically, it adjusts the vehicle's throttle and brake based on the following factors: the vehicle's average throttle and brake ${meanThrottle}$, ${meanBrake}$ in the past 5 steps, ${currentThrottle}$ and ${currentBrake}$ at this time step returned by the original controller of the behavior agent, and the worst-case robustness degrees of predicted acceleration flowpipes. It also sets the upper bounds on the throttle and brake, which are represented as ${maxThrottle}$, ${maxBrake}$, and takes the ${currentSpeed}$ at this step and a lower threshold of speed ${minSpeed}$ into consideration. When ${currentSpeed}$ is over the ${minSpeed}$, and the predicted flowpipe of acceleration satisfies the requirement at a given level ($\underline{\rho}(\varphi,\omega,t) > \rho_{thre}$), the controller simply constrains the brake and throttle under the maximum values. If the prediction contains only a hard brake (${ViolationOnDeceleration}$), the controller uses $\underline{\rho}(\varphi,\omega,t)$ with a correction parameter $\rho_{corr}$ to calculate an adjusted parameter, and applies to the average brake to obtain a ${newBrake}$, and the final brake in this case will be the lower one between ${newBrake}$ and ${maxBrake}$. When the prediction only contains a sharp acceleration (${ViolationOnAcceleration}$), the controller adjusts the throttle in a similar way as in the ${ViolationOnDeceleration}$ case. If there are both a hard brake and a sharp acceleration predicted in the future trace (${ViolationOnBoth}$), the controller chooses to adapt the brake, since if the hard brake will happen first, it needs to smooth the brake, and if the sharp acceleration will happen first, then conducting a moderate brake will help to reduce the acceleration and avoid a hard brake later. And if ${currentSpeed}$ is lower than the ${minSpeed}$, the controller uses $\underline{\rho}(\varphi,\omega,t)$ with $\rho_{corr}$ to adjust the throttle to avoid speed being too low on the road. The final brake and throttle will be ensured to be positive before being returned to the vehicle. Applying $\underline{\rho}(\varphi,\omega,t)$ with $\rho_{corr}$ as an adjustment parameter enables the controller to take the degree of violation/the closeness of possible violation into consideration and helps to provide a moderate throttle and brake, instead of a sharp one. 

\subsubsection{Datasets and experimental setup.}
We consider two groups of drivers categorized based on their driving behaviors and habits: cautious and aggressive, which are simulated with corresponding behavior agents in the CARLA simulator. 
For each behavior type, it is simulated with 15 different seeds, and 50 episodes for each seed, which makes it 750 traces in total. Each time step in the simulation represents 0.1s. We use 80\% for training, 10\% for testing, and 10\% for validation. We segment the data into samples of 50 steps by sliding windows. In the cautious and aggressive driver dataset, 82.6\% and 82.3\% of segments exhibit violations, respectively.

We train two different LSTM models for cautious and aggressive behavior agents, which aims at predicting the acceleration in the future 20 steps using the historical data of the past 30 steps. The input features include the ego vehicle's velocity, acceleration, lateral distance to detected obstacles, yaw rate, relative distance to the leading vehicle, relative speed to the leading vehicle, ego vehicle's throttle, steer, brake, as well as the change in acceleration compared to the last step. The LSTM for the cautious behavior agent is trained for 100 epochs and the LSTM for the aggressive agent is trained for 250 epochs. Given a chosen SRT and dropout rate, we repeat Bayesian LSTM predictions for 30 times using the Monte Carlo method. We estimate Gaussian distributions using these predictions and obtain flowpipes under a 95\% confidence level.
We use the validation datasets to select the optimal SRT and dropout rate under the guidance of the proposed loss function, and use the selected optimal SRT and dropout rate to generate Bayesian LSTM predictions for the testing datasets.
Finally, we apply the proposed approach in a closed-loop simulation. For each behavior type, the simulation runs with 4 different seeds and 30 episodes for each seed.
We set ${minThrottle}=0.4$, ${maxThrottle}=0.6$, ${minBrake}=0.4$, ${maxBrake}=0.6$, ${minSpeed}=5$, $\rho_{thre}=-3.0$, $\rho_{corr}=-3.0$.

\subsubsection{Experimental results.}
Similar to the diabetic case study, we compare the performance of the proposed STL-U quantitative predictive monitor with a baseline predictive monitor that does not account for the uncertainty, both using the same Bayesian LSTM models equipped with the optimal SRTs and dropout rates. 
\tabref{tab:rq2_driving_case} shows that, compared to the baseline, the proposed STL-U quantitative predictive monitor leads to earlier detection of impending hazards and higher F1 scores of requirement satisfaction. 

\tabref{tab:rq3_driving_case} shows the results of closed-loop simulation, which demonstrate that the proposed predictive monitoring and control approach improves the safety of car-following behavior by increasing the car-following distance, decreasing the number of safety hazards, while maintaining comparable vehicle speed.

\begin{table}
{%
\centering
\begin{tabular}{|l|c|cc|cc|}
\hline
& & \multicolumn{2}{|c|}{\textbf{Baseline}} & \multicolumn{2}{|c|}{\textbf{Proposed}} \\
\textbf{Model} & \textbf{Hazard} & \textbf{Time} & \textbf{F1} & \textbf{Time} & \textbf{F1} \\
\hline
\multirow{3}{*}{Cautious} & High & 0.30  & 0.17 & \textbf{1.87} & \textbf{0.82} \\
                        & Low & 0.09 & 0.01 & \textbf{1.68} & \textbf{0.69} \\
                        & Total & 0.13 & 0.15 & \textbf{1.70} & \textbf{0.87} \\
\hline
\multirow{3}{*}{Aggressive} & High & 0.56 & 0.45 & \textbf{1.69} & \textbf{0.86} \\ 
                        & Low & 0.45 & 0.10 & \textbf{1.50} & \textbf{0.79} \\
                        & Total & 0.45 & 0.41 & \textbf{1.50} & \textbf{0.90} \\ 
\hline
\end{tabular}}
\caption{Comparing the baseline and the proposed predictive monitors in terms of average pre-alert time (s) and F1 scores of requirement satisfaction. (High: acceleration $\geq 6m/s^2$, Low: acceleration $\leq -6m/s^2$)}\label{tab:rq2_driving_case}
\end{table}

\begin{table}
\setlength{\tabcolsep}{1mm}
{%
\centering
\begin{tabular}{|l|l|c|c|}
\hline
\textbf{Model} & \textbf{Metric} & \textbf{Baseline} & \textbf{Proposed}  \\
\hline
\multirow{3}{*}{Cautious} & Average hazard (High) & 1.26 & \textbf{0.10} \\
                        & Average hazard (Low) & 3.62 & \textbf{0.41} \\
                        & Average hazard (Total) & 4.88 & \textbf{0.51} \\
                        & Average distance (m) & 14.52 & \textbf{19.98} \\
                        & Average speed (m/s) & \textbf{4.46} & 4.40 \\ 
\hline 
\multirow{3}{*}{Aggressive} & Average hazard (High) & 0.71 & \textbf{0.03} \\
                        & Average hazard (Low) & 3.47 & \textbf{0.49} \\
                        & Average hazard (Total) & 4.18 & \textbf{0.52} \\
                        & Average distance (m) & 12.62 & \textbf{20.33} \\
                        & Average speed (m/s) & \textbf{4.67} & 4.36 \\ 
\hline
\end{tabular}}
\caption{Comparing the baseline controller and the proposed approach's performance in closed-loop simulation.}\label{tab:rq3_driving_case}
\end{table}



\begin{thebibliography}{27}
\providecommand{\natexlab}[1]{#1}

\bibitem[{American\_Diabetes\_Association(2022)}]{care2022care}
American\_Diabetes\_Association. 2022.
\newblock Standards of Medical Care in Diabetes—2022.
\newblock \emph{Diabetes Care}, 45: S17.

\bibitem[{Baharisangari et~al.(2021)Baharisangari, Gaglione, Neider, Topcu, and Xu}]{baharisangari2021uncertainty}
Baharisangari, N.; Gaglione, J.-R.; Neider, D.; Topcu, U.; and Xu, Z. 2021.
\newblock Uncertainty-Aware Signal Temporal Logic Inference.
\newblock In \emph{Software Verification}, 61--85. Springer.

\bibitem[{Banks, Plant, and Stanton(2018)}]{banks2018driver}
Banks, V.~A.; Plant, K.~L.; and Stanton, N.~A. 2018.
\newblock Driver error or designer error: Using the Perceptual Cycle Model to explore the circumstances surrounding the fatal Tesla crash on 7th May 2016.
\newblock \emph{Safety science}, 108: 278--285.

\bibitem[{Bartocci et~al.(2018)Bartocci, Deshmukh, Donz{\'e}, Fainekos, Maler, Ni{\v{c}}kovi{\'c}, and Sankaranarayanan}]{bartocci2018specification}
Bartocci, E.; Deshmukh, J.; Donz{\'e}, A.; Fainekos, G.; Maler, O.; Ni{\v{c}}kovi{\'c}, D.; and Sankaranarayanan, S. 2018.
\newblock Specification-based monitoring of cyber-physical systems: a survey on theory, tools and applications.
\newblock In \emph{Lectures on Runtime Verification}, 135--175. Springer.

\bibitem[{Bengio, Goodfellow, and Courville(2017)}]{bengio2017deep}
Bengio, Y.; Goodfellow, I.; and Courville, A. 2017.
\newblock \emph{Deep learning}.
\newblock MIT press Cambridge, MA, USA.

\bibitem[{Bolton, Bass, and Siminiceanu(2013)}]{bolton2013using}
Bolton, M.~L.; Bass, E.~J.; and Siminiceanu, R.~I. 2013.
\newblock Using formal verification to evaluate human-automation interaction: A review.
\newblock \emph{IEEE Transactions on Systems, Man, and Cybernetics: Systems}, 43(3): 488--503.

\bibitem[{CARLATeam(2023)}]{CARLA}
CARLATeam. 2023.
\newblock CARLA 0.9.13 Release.
\newblock \url{https://carla.org/2021/11/16/release-0.9.13/}.

\bibitem[{Fainekos and Pappas(2009)}]{fainekos2009robustness}
Fainekos, G.~E.; and Pappas, G.~J. 2009.
\newblock Robustness of temporal logic specifications for continuous-time signals.
\newblock \emph{Theoretical Computer Science}, 410(42): 4262--4291.

\bibitem[{Feng et~al.(2016)Feng, Wiltsche, Humphrey, and Topcu}]{feng2016synthesis}
Feng, L.; Wiltsche, C.; Humphrey, L.; and Topcu, U. 2016.
\newblock Synthesis of human-in-the-loop control protocols for autonomous systems.
\newblock \emph{IEEE Transactions on Automation Science and Engineering}, 13(2): 450--462.

\bibitem[{Gal(2016)}]{gal2016uncertainty}
Gal, Y. 2016.
\newblock \emph{Uncertainty in deep learning}.
\newblock Ph.D. thesis, University of Cambridge.

\bibitem[{Hochreiter and Schmidhuber(1997)}]{hochreiter1997long}
Hochreiter, S.; and Schmidhuber, J. 1997.
\newblock Long short-term memory.
\newblock \emph{Neural computation}, 9(8): 1735--1780.

\bibitem[{Kovatchev et~al.(2009)Kovatchev, Breton, Dalla~Man, and Cobelli}]{kovatchev2009biosimulation}
Kovatchev, B.~P.; Breton, M.; Dalla~Man, C.; and Cobelli, C. 2009.
\newblock Biosimulation Modeling for Diabetes: In Silico Preclinical Trials: A Proof of Concept in Closed-Loop Control of Type 1 Diabetes.
\newblock \emph{Journal of diabetes science and technology (Online)}, 3(1): 44.

\bibitem[{Lindemann et~al.(2023)Lindemann, Qin, Deshmukh, and Pappas}]{lindemann2023conformal}
Lindemann, L.; Qin, X.; Deshmukh, J.~V.; and Pappas, G.~J. 2023.
\newblock Conformal prediction for STL runtime verification.
\newblock In \emph{Proceedings of the ACM/IEEE 14th International Conference on Cyber-Physical Systems (with CPS-IoT Week 2023)}, 142--153.

\bibitem[{Luckcuck et~al.(2019)Luckcuck, Farrell, Dennis, Dixon, and Fisher}]{luckcuck2019formal}
Luckcuck, M.; Farrell, M.; Dennis, L.~A.; Dixon, C.; and Fisher, M. 2019.
\newblock Formal specification and verification of autonomous robotic systems: A survey.
\newblock \emph{ACM Computing Surveys (CSUR)}, 52(5): 1--41.

\bibitem[{Lum et~al.(2021)Lum, Bailey, Barnes-Lomen, Naranjo, Hood, Lal, Arbiter, Brown, DeSalvo, Pettus et~al.}]{lum2021real}
Lum, J.~W.; Bailey, R.~J.; Barnes-Lomen, V.; Naranjo, D.; Hood, K.~K.; Lal, R.~A.; Arbiter, B.; Brown, A.~S.; DeSalvo, D.~J.; Pettus, J.; et~al. 2021.
\newblock A real-world prospective study of the safety and effectiveness of the loop open source automated insulin delivery system.
\newblock \emph{Diabetes technology \& therapeutics}, 23(5): 367--375.

\bibitem[{Ma et~al.(2020)Ma, Gao, Feng, and Stankovic}]{ma2020stlnet}
Ma, M.; Gao, J.; Feng, L.; and Stankovic, J. 2020.
\newblock STLnet: Signal temporal logic enforced multivariate recurrent neural networks.
\newblock \emph{Advances in Neural Information Processing Systems}, 33: 14604--14614.

\bibitem[{Ma et~al.(2021)Ma, Stankovic, Bartocci, and Feng}]{ma2021predictive}
Ma, M.; Stankovic, J.; Bartocci, E.; and Feng, L. 2021.
\newblock Predictive monitoring with logic-calibrated uncertainty for cyber-physical systems.
\newblock \emph{ACM Transactions on Embedded Computing Systems (TECS)}, 20(5s): 1--25.

\bibitem[{Ma, Stankovic, and Feng(2021)}]{ma2021toward}
Ma, M.; Stankovic, J.~A.; and Feng, L. 2021.
\newblock Toward formal methods for smart cities.
\newblock \emph{Computer}, 54(9): 39--48.

\bibitem[{Maler and Nickovic(2004)}]{maler2004monitoring}
Maler, O.; and Nickovic, D. 2004.
\newblock Monitoring temporal properties of continuous signals.
\newblock In \emph{Formal Techniques, Modelling and Analysis of Timed and Fault-Tolerant Systems}, 152--166. Springer.

\bibitem[{Man et~al.(2014)Man, Micheletto, Lv, Breton, Kovatchev, and Cobelli}]{man2014uva}
Man, C.~D.; Micheletto, F.; Lv, D.; Breton, M.; Kovatchev, B.; and Cobelli, C. 2014.
\newblock The UVA/PADOVA type 1 diabetes simulator: new features.
\newblock \emph{Journal of diabetes science and technology}, 8(1): 26--34.

\bibitem[{Qin and Deshmukh(2020)}]{qin2020clairvoyant}
Qin, X.; and Deshmukh, J.~V. 2020.
\newblock Clairvoyant Monitoring for Signal Temporal Logic.
\newblock In \emph{International Conference on Formal Modeling and Analysis of Timed Systems}, 178--195. Springer.

\bibitem[{SafeBenchTeam(2023)}]{SafeBench}
SafeBenchTeam. 2023.
\newblock SafeBench: A Benchmark for Evaluating Autonomous Vehicles in Safety-critical Scenarios.
\newblock \url{https://safebench.github.io/}.

\bibitem[{Sharp, Preece, and Rogers(2019)}]{sharp2019interaction}
Sharp, H.; Preece, J.; and Rogers, Y. 2019.
\newblock \emph{Interaction design: beyond human-computer interaction}.
\newblock NY: Wiley.

\bibitem[{Slattery, Amiel, and Choudhary(2018)}]{slattery2018optimal}
Slattery, D.; Amiel, S.; and Choudhary, P. 2018.
\newblock Optimal prandial timing of bolus insulin in diabetes management: a review.
\newblock \emph{Diabetic Medicine}, 35(3): 306--316.

\bibitem[{Visconti et~al.(2021)Visconti, Bartocci, Loreti, and Nenzi}]{visconti2021online}
Visconti, E.; Bartocci, E.; Loreti, M.; and Nenzi, L. 2021.
\newblock Online monitoring of spatio-temporal properties for imprecise signals.
\newblock In \emph{Proceedings of the 19th ACM-IEEE International Conference on Formal Methods and Models for System Design}, 78--88.

\bibitem[{Yang et~al.(2022)Yang, Zhong, Feng, Li, Shao, and Liu}]{yang2022robot}
Yang, S.; Zhong, Y.; Feng, D.; Li, R. Y.~M.; Shao, X.-F.; and Liu, W. 2022.
\newblock Robot application and occupational injuries: are robots necessarily safer?
\newblock \emph{Safety science}, 147: 105623.

\bibitem[{Yoon and Sankaranarayanan(2021)}]{yoon2021predictive}
Yoon, H.; and Sankaranarayanan, S. 2021.
\newblock Predictive runtime monitoring for mobile robots using logic-based bayesian intent inference.
\newblock In \emph{2021 IEEE International Conference on Robotics and Automation (ICRA)}, 8565--8571. IEEE.

\end{thebibliography}
\end{document}